\documentclass[10pt,twocolumn,letterpaper]{article}

\usepackage{iccv}
\usepackage{times}
\usepackage{epsfig}
\usepackage{graphicx}
\usepackage{amsmath}
\usepackage{amssymb}

\usepackage{multirow}
\usepackage{booktabs}
\usepackage[breaklinks=true,bookmarks=false]{hyperref}

\iccvfinalcopy 

\def\iccvPaperID{5} 

\ificcvfinal\fi

\begin{document}

\title{EENA: Efficient Evolution of Neural Architecture}

\author{Hui Zhu\textsuperscript{\rm 1,2}, Zhulin An\textsuperscript{\rm 1}\thanks{Corresponding author of this work}, Chuanguang Yang\textsuperscript{\rm 1}, Kaiqiang Xu\textsuperscript{\rm 1}, Erhu Zhao\textsuperscript{\rm 1}, Yongjun Xu\textsuperscript{\rm 1}\\
\textsuperscript{\rm 1}Institute of Computing Technology, Chinese Academy of Sciences\\
\textsuperscript{\rm 2}University of Chinese Academy of Sciences\\
{\tt\small \{zhuhui,anzhulin,yangchuanguang,xukaiqiang,zhaoerhu,xyj\}@ict.ac.cn}
}

\maketitle
\ificcvfinal\thispagestyle{empty}\fi

\begin{abstract}
   Latest algorithms for automatic neural architecture search perform remarkable but are basically directionless in search space and computational expensive in the training of every intermediate architecture. In this paper, we propose a method for efficient architecture search called EENA (Efficient Evolution of Neural Architecture). Due to the elaborately designed mutation and crossover operations, the evolution process can be guided by the information have already been learned. Therefore, less computational effort will be required while the searching and training time can be reduced significantly. On CIFAR-10 classification, EENA using minimal computational resources (0.65 GPU$-$days) can design highly effective neural architecture which achieves 2.56\% test error with 8.47M parameters. Furthermore, the best architecture discovered is also transferable for CIFAR-100.
\end{abstract}

\section{Introduction}
Convolutional Neural Network has a prominent performance in computer vision, object detection and other fields by extracting features through neural architectures which imitate the mechanism of human brain. Human-designed neural architectures such as ResNet  \cite{resnet}, DenseNet \cite{densenet}, PyramidNet \cite{pyramid} and so on which contain several effective blocks are successively proposed to increase the accuracy of image classification. In order to design neural architectures adaptable for various datasets, more researchers have a growing interest in studying the algorithmic solutions based on human experience to achieve automatic neural architecture search \cite{nas,hierarchical,darts,enas,pnas,proxylessnas,efficientnet}.

Many architecture search algorithms perform remarkable but demand for lots of computational effort. For example, obtaining a state-of-the-art architecture for CIFAR-10 required 7 days with 450 GPUs of evolutionary algorithm \cite{AmoebaNet} or used 800 GPUs for 28 days of reinforcement learning \cite{nas}. The latest algorithms based on reinforcement learning (RL) \cite{enas}, sequential model-based optimization (SMBO) \cite{pnasreal} and Bayesian optimization \cite{bayesian} over a discrete space are proposed to speed up the search process but the basically directionless search leads to a large number of architectures evaluations required. Although several algorithms based on gradient descent over a continuous space, such as DARTS \cite{darts} and NAO \cite{naonet} address this problem to some extent, the training of every intermediate architecture is still computational expensive. 

In this work, we propose a method for efficient architecture search called EENA (Efficient Evolution of Neural Architecture) guided by the experience gained in the prior learning to speed up the search process and thus consume less computational effort. The concept, guidance of experience gained, is inspired by Net2Net \cite{net2net}, which generate large networks by transforming small networks via function-preserving. There are several precedents \cite{rlnt,ibm,eas} based on this for neural architecture search, but the basic operations are limited to the experience in parameters and are relatively simple, so the algorithms may degenerate into a random search. We absorb more basic blocks of classical networks, discard several ineffective blocks and even extend the guidance of experience gained to the prior architectures by crossover in our method. Due to the loss continue to decrease and the evolution becomes directional, robust and globally optimal models can be discovered rapidly in the search space.

Our experiments (Sect.~\ref{sec4}) of neural architecture search on CIFAR-10 show that our method using minimal computational resources (0.65 GPU$-$days\footnotemark \footnotetext{All of our experiments were performed using a NVIDIA Titan Xp GPU.}) can design highly effective neural architecture that achieves 2.56\% test error with 8.47M parameters. We further transfer the best architecture discovered on CIFAR-10 to CIFAR-100 datasets and the results perform remarkable as well. 

Our contributions are summarized as follows:

\noindent(1) We are the first to propose the crossover operation guided by experience gained to effectively reuse the prior learned architectures and parameters.

\noindent(2) We study a large number of basic mutation operations absorbed from typical architectures and select the ones that have significant effects.

\noindent(3) We achieve remarkable architecture search efficiency (2.56\% error on CIFAR-10 in 0.65 GPU-days) which we attribute to the use of EENA.

\noindent(4) We show that the neural architectures searched by EENA on CIFAR-10 are transferable for CIFAR-100 datasets.

Part of the code implementation and several models we searched on CIFAR-10 of EENA is available at \url{https://github.com/ICCV-5-EENA/EENA}.

\section{Related Work}
\label{sec2}
In this section, we review human-designed neural architectures and automatic neural architecture search which are most related to this work.

\paragraph{Human-Designed Neural Architectures.} Since convolutional neural networks were first used to deal with the problems in the field of computer vision, human-designed neural architectures constantly improve the classification accuracy on specific datasets. As the crucial factor affecting the performance of neural network, many excellent neural architectures have been designed. Chollet \etal \cite{xception} propose to replace Inception with depthwise separable convolutions to reduce the number of parameters. Grouped convolutions given by Krizhevsky \etal \cite{alexnet} is used to distribute the model over two GPUs and Xie \etal \cite{resnext} further propose that increasing cardinality is more effective than going deeper or wider based on this. He \etal \cite{resnet} solve the degradation problem of deep neural networks by residual blocks. Huang \etal \cite{densenet} propose dense blocks to solve the vanishing-gradient problem and substantially reduce the number of parameters. Hu \etal \cite{senet} propose the Squeeze-and-Excitation block that adaptively recalibrates channel-wise feature responses by explicitly modelling interdependencies between channels. In addition to the improvements on convolutional layer, many other methods have been proposed to further optimize the network. Lin \etal \cite{gap} utilize global average pooling to replace the traditional fully connected layers to enforce correspondences between feature maps and categories and avoid overfitting. Some regularization methods, such as dropout \cite{dropout}, dropblock \cite{dropblock} and shake-shake \cite{shakeshake} and so on, also improve the generalization ability of neural networks. Human-designed neural architectures rely heavily on expert experience, and it will be extremely difficult to design the most suitable neural architecture when faced with a new task for image. In this work, we fully draw on the excellent experience of predecessors and design an efficient neural architecture search method.

\paragraph{Automatic Neural Architecture Search.} Many different search strategies have been proposed to explore the space of neural architectures, including random search, evolutionary algorithm (EA), reinforcement learning (RL), Bayesian optimization (BO) and gradient-based methods. Early approaches \cite{neat,HyperNEAT} use evolutionary algorithms to optimize both the neural architecture and its weights. However, faced with the scale of contemporary neural architectures with millions of weights, recent approaches based on evolutionary algorithm have been imporved in some ways, such as using gradient-based methods for optimizing weights \cite{large,AmoebaNet,hierarchical} or modifying the architecture by network morphisms \cite{morphism,net2net,eas}. The generation of a neural architecture can be regarded as the agent's action, thus, many approaches based on reinforcement learning \cite{nas,nasnet} have been proposed to deal with neural architecture search. Kandasamy \etal \cite{bayesian} derive kernel functions for architecture search spaces in order to use classic GP-based BO method, but compared with others, BO method shows no obvious advantages. In contrast to the gradient-free  methods above, as gradient-based approaches, Liu \etal \cite{darts} propose a continuous relaxation of the search space and Luo \etal \cite{naonet} use an encoder network mapping neural network architectures into a continuous space. We can notice that the algorithms of automatic neural architecture search pay more and more attention to the efficiency of algorithms (such as search time) besides focusing on the effect of neural architectures discovered. In this work, we propose a different method based on evolutionary algorithm and network morphisms for efficient architecture search and achieve remarkable results on the classification datasets.

\section{Proposed Methods}
In this section, we illustrate our basic mutation and crossover operations with an example of several connected layers which come from a simple convolutional neural network and describe the method of selection and discard of individuals from the population in the evolution process. 

\begin{figure*}[t]
  \centering
  \includegraphics[width=0.92\textwidth]{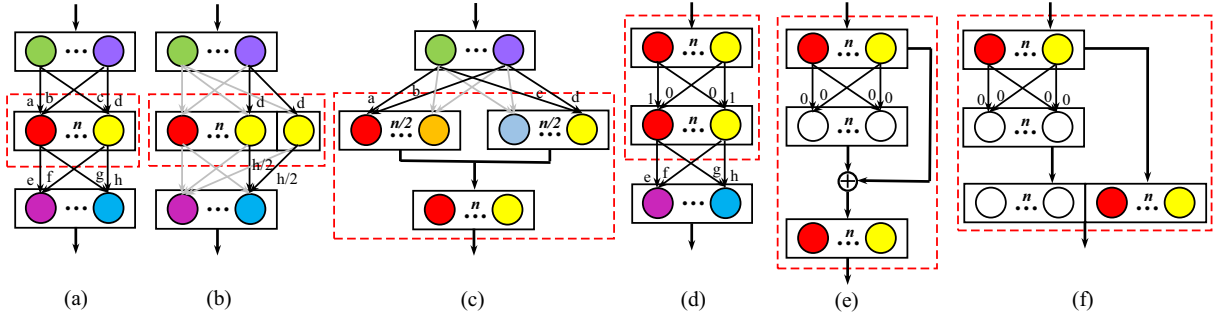}
  \caption{Visualization of the teacher network (a) and several mutation operations (b$\sim$f). The rectangles and circles represent the convolutional layers and feature maps or filters, respectively. The same color means identical and white means 0 value. The parts in the red dashed box are equivalent.}
  \label{fig1}
\end{figure*}

\subsection{Search Space and Mutation Operations}
\label{sec3.1}
As we mentioned in the Sect.~\ref{sec2}, the birth of a better network architecture is usually achieved based on the local improvements. A good design of automatic neural architecture search should be based on a large number of excellent human-designed architctures. In addition, some of the existing methods \cite{net2net,ibm} based on function-preserving are briefly reviewed in this section and our method is built on them. Specifically, we absorb more blocks of classical networks such as dense block, add some effective changes such as noises for new parameters and discard several ineffective operations such as kernel widening in our method.

Our method explores the search space by mutation and crossover operations and every mutation operation refers to a random change for an individual. $x$ is the input to the network, the guidance of experience gained in parameters is to choose a new set of parameters $\theta^{'}$ for a student network $g(x;\theta^{'})$ which transform from the teacher network $h(x;\theta)$ such that:
\begin{equation}
\forall x:h(x;\theta)=g(x;\theta^{'})\footnotemark \footnotetext{The '$=$' here doesn't mean completely equivalent, noise may be added to make the student more robust.}. 
\end{equation}
Assume that the $i$-th convolutional layer to be changed is represented by a $(k_{1},k_{2}, c, f)$  shaped matrix $W^{(i)}$. The input for the convolution operation in layer $i$ is represented as $X^{(i)}$ and the processing of BatchNorm and ReLU is expressed as $\varphi$. In this work, we consider the following mutation operations.

\paragraph{Widen a Layer.} Fig.~\ref{fig1}(b) is an example of this operation. $W^{(i)}$ is extend by replicating the parameters along the last axis at random and the parameters in $W^{(i+1)}$ need to be divided along the third axis corresponding to the counts of the same filters in the $i$-th layer. $U$ is the new parameter matrix and $f^{'}$ is the number of filters in the layer $i$+1. Specifically, A noise $\delta$ is randomly added to every new parameter in $W^{(i+1)}$ to break symmetry.
\begin{equation}
g(j)=
\left\{  
	\begin{array}{lr}
	j  & j \leq f \\ 
	random \, sample \, from \lbrace 1,2,\cdots,f \rbrace  & j > f
	\end{array}
\right.,
\end{equation}
\begin{equation}
U_{k_{1},k_{2},c,j}^{(i)} = W_{k_{1},k_{2},c,g_{(j)}}^{(i)},
\end{equation}
\begin{equation}
U_{k_{1},k_{2},j,f^{'}}^{(i+1)} = \frac{ W_{k_{1},k_{2},g(j),f^{'}}^{(i+1)}}{card({x \vert g(x)=g(j)})} \cdot {(1+ \delta)}, \, \delta \in \left[0, 0.05 \right].
\end{equation}

\paragraph{Branch a Layer.} Fig.~\ref{fig1}(c) is an example of this operation. This operation adds no further parameters and will always be combined with other operations. $U$ and $V$ are the new parameter matrices which can be expressed as:
\begin{equation}
\begin{split}
U_{k_{1},k_{2},c,j}^{(i)} = W_{k_{1},k_{2},c,m}^{(i)} \quad m \in \left[0,\lfloor \frac{f}{2} \rfloor \right],\\
V_{k_{1},k_{2},c,l}^{(i)} = W_{k_{1},k_{2},c,m}^{(i)} \quad m \in \left(\lfloor \frac{f}{2} \rfloor,f \right].
\end{split}
\end{equation}
The convolutional layer will be reformulated as:
\begin{equation}
Concat \left( \varphi \left( X^{(i)} \cdot U_{k_{1},k_{2},c,j}^{(i)} \right), \varphi \left( X^{(i)} \cdot V_{k_{1},k_{2},c,l}^{(i)} \right)\right).
\end{equation}

\paragraph{Insert a Single Layer.} Fig.~\ref{fig1}(d) is an example of this operation. The new layer weight matrix $U^{(i+1)}$ with a $k_{1} \times k_{2}$ kernel is initialized to an identity matrix. $ReLU(x) = max\{x,0\}$ satisfies the restriction for the activation function $\sigma$:
\begin{equation}
\forall x:\sigma(x)=\sigma \left( I\sigma(x) \right),
\end{equation}
so this operation is possible and the new matrix can be expressed as:
\begin{equation}
U_{j,l,a,b}^{(i+1)}=
\left\{  
	\begin{array}{lr}
	1  & j= \frac{k_{1}+1}{2} \wedge l= \frac{k_{2}+1}{2} \wedge a=b \\ 
	0  & \rm otherwise
	\end{array}
\right. .
\end{equation}

\paragraph{Insert a Layer with Shortcut Connection.} Fig.~\ref{fig1}(e) is an example of this operation. All the parameters of the new layer weight matrix $U^{(i+1)}$ are initialized to 0.	
The convolutional layer will be reformulated as:
\begin{equation}
Add \left( \varphi \left( X^{(i+1)} \right), \varphi \left( X^{(i+1)} \cdot U^{(i+1)} \right) \right) .
\end{equation}

\paragraph{Insert a Layer With Dense Connection.} Fig.~\ref{fig1}(f) is an example of this operation. All the parameters of the new layer weight matrix $U^{(i+1)} $ are initialized to 0.
The convolutional layer will be reformulated as:
\begin{equation}
Concat \left( \varphi \left( X^{(i+1)} \cdot U^{(i+1)} \right), \varphi \left( X^{(i+1)} \right)\right) .
\end{equation}

In addition, many other important methods, such as separable convolution, grouped convolution and bottleneck etc. can be absorbed into the mutation operations. For these, we run several simple tests and notice that the search space is expanded but the accuracy of classification is not improved. Therefore, we finally abandoned these operations in our experiment.

\subsection{Crossover Operation}
\label{sec3.2}
Crossover refers to the combination of the prominent parents to produce offsprings which may perform even more excellent. The parents refer to the architectures with high fitness (accuracy) that have been already discovered and every offspring can be considered as a new exploration of the search space. Obviously, although our mutation operations reduce the computational effort of the repeated retraining, the exploration of the search space is still random without taking advantage of the experience already gained in prior architectures. It is crucial and difficult to find a crossover operation that can effectively reuse the parameters already trained and even produce the next generation guided by experience of the prior excellent architectures. 
\begin{figure}[t]
  \begin{center}
    \includegraphics[width=0.45\textwidth]{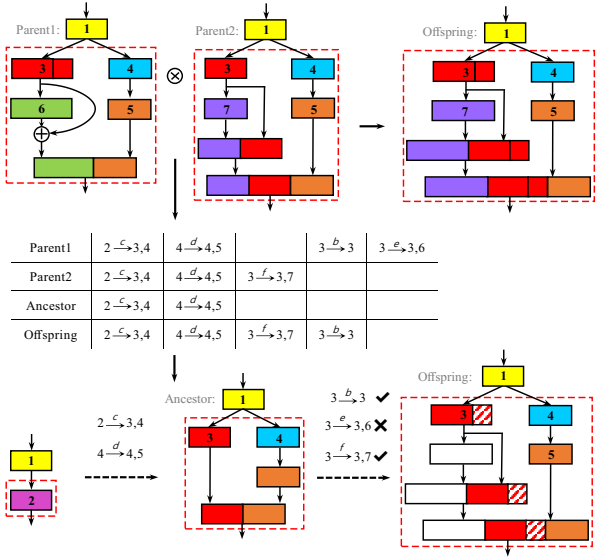}
  \end{center}
  \caption{Visualization of the process that the parents produce an offspring by crossover operation. b$\sim$f correspond to the mutation operations in Fig.~\ref{fig1} and 1$\sim$7 represent the serial number of layers. The same colored rectangles represent the identical layers and white means 0 value. The parts in the red dashed box are equivalent.}
\label{fig2}
\end{figure}

NEAT \cite{neat}, as a existing method in the field of evolutionary algorithm, identify which genes line up with which by assigning the innovation number to each node gene. However, this method is limited to the fine-grained crossover for nodes and connections, and will destroy the parameters that have already been trained. 

We notice that the architectures with high fitness all derive from the same ancestor of some point in the past (At worst, the ancestor is the initial architecture). Whenever a new architecture appears (through mutation operations), we record the type and the location of the mutation operation. Based on these, we can track the historical origins and find the common ancestor of the two individuals with high fitness. Then the offsprings inherit the same architecture (ancestor) and randomly inherit the different parts of architectures of the parents.

Fig.~\ref{fig2} is a visual example of the crossover operation in our experiments. Based on the records about the previous mutation operations for each individual (for Parent1, mutation c, d, b, e occurred at layer 2, 4, 3, 3, respectively and for Parent2, mutation c, d, f occurred at layer 2, 4, 3, respectively), the common ancestor of the parents (Ancestor with mutation c, d occurred at layer 2, 4) can be easily found. The mutation operations of the two parents different from each other are selected and added to the ancestor architecture according to a certain probability by the mutation operations (mutation b, f occurred at layer 3, 3 are inherited by Offspring and mutation e is randomly discarded). 

\subsection{The Selection and Discard of Individuals in Evolutionary Algorithm}
\label{sec3.3}
\paragraph{The Selection of Individuals.} Our evolutionary algorithm uses tournament selection \cite{tournament} to select an individual for mutation: a fraction $k$ of individuals is selected from the population randomly and the individual with highest fitness is final selected from this set. For crossover, the two individuals with the highest fitness but different architectures will be selected. 
\paragraph{The Discard of Individuals.} In order to constrain the size of the population, the discard of individuals will be accompanied by the generation of each new individual when the population size reaches $N$. We regulate aging and non-aging evolutions \cite{AmoebaNet} via a variable $\lambda$ to affect the convergence rate and overfit: Discarding the worst model with probability $\lambda$ and the oldest model with $1-\lambda$ within each round.

\section{Experiments}
\begin{figure*}[t]
  \centering
  \includegraphics[width=0.92\linewidth]{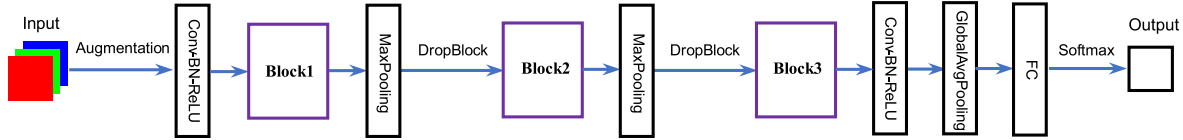}
  \caption{The initial model designed in our experiments.}
  \label{figure3}
\end{figure*}

\begin{figure*}[t]
  \centering
  \includegraphics[width=0.92\linewidth]{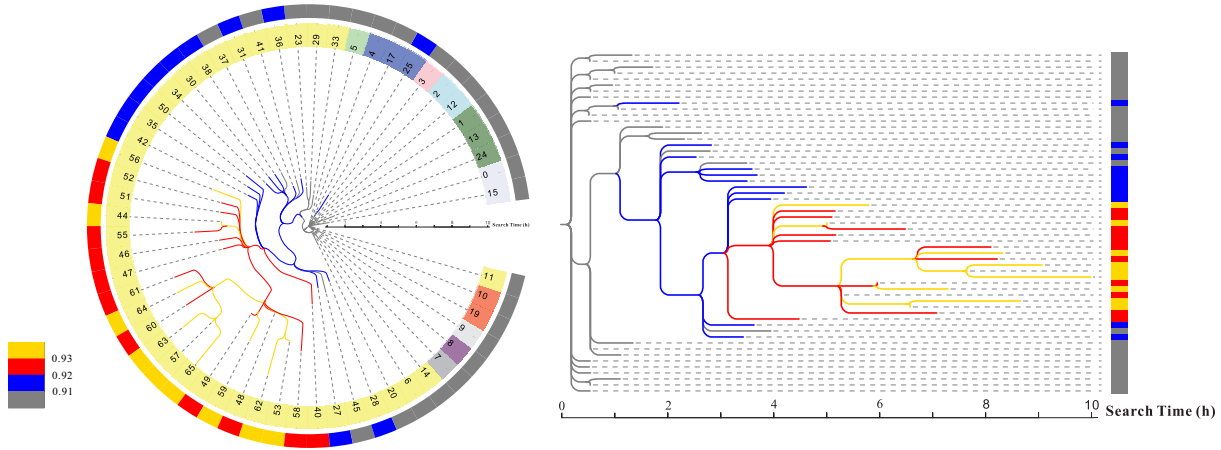}
  \caption{The phylogenetic tree visualized one search process on CIFAR-10. In the circle phylogenetic tree, the color of the outermost circle represents fitness, and the color of the penultimate circle represents ancestor. In the rectangular phylogenetic tree, the color on the right side represents fitness. From the inside to the outside in the left figure and from left to right in the right figure, along the direction of time axis, the connections represent the relationship from ancestors to offsprings.}
\label{figure4}
\end{figure*}

\label{sec4}
In this section, we report the performances of EENA in neural architecture search on CIFAR-10 and the feasibility of transferring the best architecture discovered on CIFAR-10 to CIFAR-100. In addition, we statistically analyze the effect of mutation and crossover operations. In our experiments, we start the evolution from initializing a simple convolutional neural network to show the efficiency of EENA and we use the methods of selection and discard mentioned in Sect.~\ref{sec3.3} to select individuals from the population and the mutation Sect.~\ref{sec3.1} and crossover in Sect.~\ref{sec3.2} operations to improve the neural architectures. 

\paragraph{Initial Model.} The initial model (the number of parameters is 0.67M) is sketched in Fig.~\ref{figure3}. It starts with one convolutional layer, followed by three evolutionary blocks and two MaxPooling layers for down-sampling which are connected alternately. Then another convolutional layer is added, followed by a GlobalAveragePooling layer and a Softmax layer for transformation from feature map to classification. Each MaxPooling layer has a stride of two and is followed by a DropBlock \cite{dropblock} layer with $keep\_prob = 0.8$ ($block\_size = 7$ for the first one and $block\_size = 5$ for the second one). Specifically, the first convolutional layer contains 64 filters and the last convolutional layer contains 256 filters. An evolutionary block is initialized with a convolutional layer with 128 filters. Every convolutional layer mentioned actually means a Conv-BatchNorm-ReLU block with a kernel size of $3\times3$. The weights are initialized as He normal distribution \cite{resnet} and the L2 regularization of $0.0001$ is applied to the weights.

\paragraph{Dataset.} We randomly sample 10,000 images by stratified sampling from the original training set to form a validation set for evaluate the fitness of the individuals while using the remaining 40,000 images for training the individuals during the evolution. We normalize the images using channel means and standard deviations for preprocessing and apply a standard data augmentation scheme (zero-padding with 4 pixels on each side to obtain a $40\times40$ pixels image, then randomly cropping it to size $32\times32$ and randomly flipping the image horizontally). 

\paragraph{Search on CIFAR-10.} The initial population consists of 12 individuals, each formed by a single mutation operation from the common initial model. During the process of evolution, Individual selection is determined by the fitness (accuracy) of the neural architecture evaluated on the validation set. In our experiments, the size of $k$ in selection of individuals is fixed to 3 and the variable $\lambda$ in discard of individuals is fixed to 0.5. We don't discard any individual at the beginning to make the population grow to the size of 20. Then we use selection and discard together, that is to say, the individual after mutation or crossover operations will be put back into the population after training and at the same time the discard of individuals will be executed. The mutation and crossover operations to improve the neural architectures are applied in the evolutionary block and any mutation operation is selected by the same probability. The crossover operation is executed every 5 rounds, for which we select the two individuals as parents with the highest fitness but different architectures among the population. All the neural architectures are trained with a batch size of 128 using SGDR \cite{SGDR} with initial learning rate $l_{max} = 0.05$, $T_{0} = 1$ and $T_{mult} = 2$. The initial model is trained for 63 epochs. Then, 15 epochs are trained after each mutation operation, one round of 7 epochs and another round of 15 epochs are trained after each crossover operation. One search process on CIFAR-10 is visualized in fig.~\ref{figure4}. In the circle phylogenetic tree of EENA, the color of the outermost circle represents fitness, and the same color of the penultimate circle represents the same ancestor. In the rectangular phylogenetic tree, the color on the right side represents fitness. From the inside to the outside in the left figure and from left to right in the right figure, along the direction of time axis, the connections represent the relationship from ancestors to offsprings. We can notice that the fitness of the population increases steadily and rapidly via mutation and crossover operations. In addition, the population is quickly taken over by a highly prominent homologous group. After the search budget is exhausted or the highest fitness of the population no longer increase over 25 rounds, the individual with highest fitness will be extracted as the best neural architecture for post-training.

\begin{table}[t]
  \caption{Comparison against state-of-the-art recognition results on CIFAR-10. Results marked with $\dagger$ are NOT trained with Cutout \cite{cutout}. The first block represents the performance of human-designed architectures. The second block represents results of various automatically designed architectures. Our method use minimal computational resources to achieve a low test error.}\smallskip \smallskip
  \label{table1}
  \centering
  \resizebox{1\columnwidth}{!}{
  \begin{tabular}{lccc}
    \toprule
    \multirow{2}{*}{\textbf{Method}}   & \textbf{Params}  & \textbf{Search Time} & \textbf{Test Error}\\
     & (Mil.) & (GPU-days) & (\%) \\
    \midrule
    DenseNet-BC \cite{densenet}$\,^{\dagger}$  & 25.6  & $-$ & 3.46    \\
    PyramidNet-Bottleneck \cite{pyramid}$\,^{\dagger}$ & 26.0  & $-$ & 3.31    \\
    ResNeXt + Shake-Shake \cite{shakeshake}$\,^{\dagger}$  & 26.2  & $-$ & 2.86    \\
    \midrule
    AmoebaNet-A \cite{AmoebaNet}  & 3.2  & 3150 & 3.34    \\
    Large-scale Evolution \cite{large}$\,^{\dagger}$  & 5.4  & 2600 & 5.4    \\
    NAS-v3 \cite{nas}  & 37.4  & 1800 & 3.65    \\
    NASNet-A \cite{nasnet}  & 3.3  & 1800 & 2.65    \\
    Hierarchical Evolution \cite{hierarchical}$\,^{\dagger}$  &15.7  & 300 & 3.75    \\
    PNAS \cite{pnasreal}$\,^{\dagger}$   & 3.2  & 225 & 3.41    \\
    Path-Level-EAS \cite{pnas}   & 14.3  & 200 & 2.30    \\
    NAONet \cite{naonet}   & 128  & 200 & 2.11    \\
    EAS \cite{eas}$\,^{\dagger}$  & 23.4  & 10 & 4.23    \\
    DARTS \cite{darts}   & 3.4  & 4 & 2.83    \\
    Neuro-Cell-based Evolution \cite{ibm}   & 7.2  & 1 & 3.58    \\
    ENAS \cite{enas}  & 4.6  & 0.45 & 2.89    \\
    NAC \cite{nac}   & 10  & 0.25 & 3.33    \\
    GDAS(FRC) \cite{GDAS} & 2.5 & 0.17 & 2.82 \\
    \midrule
    Ours & 8.47 & 0.65 & 2.56    \\
    Ours (adjust the number of channels) & 54.14 & $-$ & 2.21    \\
    \bottomrule
  \end{tabular}
  }
\end{table}

\paragraph{Post-Training of the Best Neural Architecture Obtained.} We conduct post-processing and post-training towards the best neural architecture designed by the EENA. The model is trained on the full training dataset until convergence using Cutout \cite{cutout} and Mixup \cite{mixup} whose configurations are the same as the original paper (a cutout size of $16\times16$ and $\alpha=1$ for mixup). Specifically, in order to reflect the fairness of the result for comparison, we don't use the latest method AutoAugment \cite{autoaugment} which has significant effects but hasn't been widely used yet. The neural architectures are trained with a batch size of 128 using SGD with learning rate $l = 0.1$ for 50 epochs to accelerate the convergence process. Then we used SGDR with initial learning rate $l_{max} = 0.1$, $T_{0} = 1$ and $T_{mult} = 2$ for 511 or 1023 epochs\footnotemark \footnotetext{We did not conduct extensive hyperparameter tuning due to limited computation resources.}. Finally, the error on the test dataset is reported. The comparison against state-of-the-art recognition results on CIFAR-10 is presented in Table~\ref{table1}. On CIFAR-10, Our method using minimal computational resources (0.65 GPU-days) can design highly effective neural cell that achieves 2.56\% test error with small number of parameters (8.47M). 

\paragraph{Fine-Tuning of the Best Neural Network.} In order to prove the good scalability of the network designed by EENA, we further fine-tune the best neural network by adjusting the number of the channels. We change the number of the filters in the three evolutionary blocks of the convolutional neural network to 1, 2, 4 times and use the same post-processing and post-training method which is mentioned above. The result after fine-tuning on CIFAR-10 is also presented in Table~\ref{table1}. We can notice that a better performance (2.21\% test error) can be achieved by simply widening the discovered convolutional neural network.

\begin{table}[t]
  \caption{The effect of mutation for a random selected individual. The first column represents the percentage that the fitness becomes better after mutation. The second column represents the percentage that the fitness becomes worse. The last column represents that the fitness is unchanged.}\smallskip \smallskip
  \label{table4}
  \centering
  \resizebox{0.95\columnwidth}{!}{
  \begin{tabular}{ccc}
    \toprule
    \multicolumn{3}{c}{\textbf{The Fitness after Mutation for a Random Individual}} \\
    \midrule
    \textbf{Better}   & \textbf{Worse}   & \textbf{No Change}\\
    \midrule
    118/310 (38.06\%)  & 182/310 (58.71\%)  & 10/310 (3.23\%)  \\
    \bottomrule
  \end{tabular}
  }
\end{table}

\begin{table}[t]
  \caption{The comparison of the effect of mutation and crossover. The second column represents the percentage of generating the Top1 fitness of the population from the individuals of Top2 fitness after mutation or crossover. The last column represents the percentage of generating the Top5 fitness of the population from the individuals of Top2 fitness after mutation or crossover.}\smallskip \smallskip
  \label{table3}
  \centering
  \resizebox{0.95\columnwidth}{!}{
  \begin{tabular}{ccc}
    \toprule
    \textbf{Operations}   & \textbf{Top1}   & \textbf{Top5}\\
    \midrule
    Mutation (Top2)  & 28/120 (23.33\%)  & 79/120 (65.83\%)  \\
    Crossover (Top2)  & 18/55 (32.73\%)  & 42/55 (76.36\%) \\
    \bottomrule
  \end{tabular}
  }
\end{table}

\paragraph{The Effect of Mutation and Crossover Operations.} We conduct several experiments and do a statistical analysis about the changes of the fitness for an individual after mutation and crossover operations. We collect the results of 310 random individuals for mutation and count the number of the offsprings with better fitness or worse fitness separately. Then we roughly estimate the effect of this operation according to the statistical results. The effect of mutation for a random selected individual is presented in Table~\ref{table4}. We also collect the results of 120 individuals with Top2 fitness for mutation and 55 pairs of individuals with Top2 fitness for crossover. Then we count the number of the offsprings with Top1 and Top5 fitness in the population separately. The comparison of the effect of mutation and crossover is presented in Table~\ref{table3}. We can notice that the mutation and crossover are both effective but crossover operation is more likely to produce individuals with high fitness (Top1 or Top5 fitness).

\paragraph{Comparison to Search Without Crossover.} Unlike random search by mutation operations, crossover as a heuristic search makes the exploration directional. In order to further verify the effect of the crossover operation, we conduct another experiment removing the crossover operation from the search process and all the other configurations remain unchanged. We run the experiment 5 times for 0.65 hours, then report a mean classification error of 3.44\% and a best classification error of 2.96\%. The result is worse than that of the original experiment with crossover. Thus, associating this result with the effect of mutation and crossover operations which is mentioned above, we confirm that the crossover operation is indeed effective.

\begin{table}[t]
  \caption{Comparison against state-of-the-art recognition results on CIFAR-100. The first block represents the performance of human-designed architectures. The second block represents the results of several automatically designed architectures.The last block represents the performance of transferring the best architecture discovered on CIFAR-10 to CIFAR-100.}\smallskip \smallskip
  \label{table2}
  \centering
  \resizebox{1\columnwidth}{!}{
  \begin{tabular}{lccc}
    \toprule
    \multirow{2}{*}{\textbf{Method}}   & \textbf{Params}   & \textbf{Search Time} & \textbf{Test Error} \\
     & (Mil.) & (GPU-days) & (\%) \\
    \midrule
    DenseNet-BC \cite{densenet}  & 25.6  & $-$ & 17.18    \\
    ResNeXt + Shake-Shake \cite{shakeshake}  & 26.2  & $-$ & 15.20    \\
    \midrule
   AmoebaNet-B \cite{AmoebaNet}  & 34.9 & 3150 & 15.80    \\
   Large-scale Evolution \cite{large}  & 40.4 & 2600 & 23.70   \\
   NASNet-A \cite{nasnet}  & 50.9 & 1800 & 16.03   \\
   PNAS \cite{pnasreal}  & 3.2 & 225 & 17.63   \\
   NAONet \cite{naonet}  & 128 & 200 & 14.75   \\
   Neuro-Cell-based Evolution \cite{ibm}  & 5.3 & 1 & 21.74   \\
   GDAS(FRC) \cite{GDAS} &2.5 & 0.17 & 18.13 \\
    \midrule
    Ours (transferred from CIFAR-10) & 8.49 & $-$ & 17.71    \\
    \bottomrule
  \end{tabular}
  }
\end{table}

\paragraph{Transfer the Best Architecture Searched on CIFAR-10 to CIFAR-100.} We further try to transfer the best architecture of highest fitness searched on CIFAR-10 to CIFAR-100 and the results perform remarkable as well. For CIFAR-100, several hyper-parameters are modified: $block\_size = 3$ for the first DropBlock layer, $block\_size = 2$ for the second and the cutout size is $8\times8$. The comparison against state-of-the-art recognition results on CIFAR-100 is presented in Table~\ref{table2}. 

\section{Conclusions and Ongoing Work}
We design an efficient method of neural architecture search based on evolution with the guidance of experience gained in the prior learning. This method takes repeatable CNN blocks (cells) as the basic units for evolution, and achieves a state-of-the-art accuracy on CIFAR-10 and others with few parameters and little search time. We notice that the initial model and the basic operations are extremely impactful to search speed and final accuracy. Therefore, we are trying to add several effective blocks such as SE block \cite{senet} as mutation operations combined with other methods that might perform effective such as macro-search \cite{macro} into our experiments.

\section*{Appendix}
\begin{figure}
  \centering
  \includegraphics[width=0.6\linewidth]{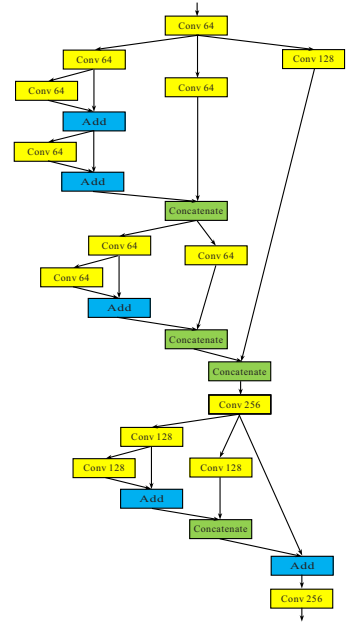}
  \caption{The best architecture discovered by EENA. 64, 128 and 256 is the number of filters.}
  \label{figure5}
\end{figure}

Here we plot the best architecture of CNN cells discovered by EENA in Fig.~\ref{figure5}.

{\small
\bibliographystyle{ieee}
\bibliography{egbib}
}

\end{document}